\documentclass[letterpaper]{article} 
\usepackage{aaai2026}  
\usepackage{amsmath}
\usepackage{amssymb}
\usepackage{amsmath}
\usepackage{times}  
\usepackage{helvet}  
\usepackage{courier}  
\usepackage[hyphens]{url}  
\usepackage{graphicx} 
\usepackage{natbib}  
\usepackage{caption} 
\usepackage{algorithm}
\usepackage{algorithmic}

\usepackage{newfloat}
\usepackage{listings}
\DeclareCaptionStyle{ruled}{labelfont=normalfont,labelsep=colon,strut=off} 
\floatstyle{ruled}
\newfloat{listing}{tb}{lst}{}
\floatname{listing}{Listing}

\usepackage{subfiles}

    \title{TSPE-GS: Probabilistic Depth Extraction for Semi-Transparent Surface Reconstruction via 3D Gaussian Splatting}

\author {
Zhiyuan Xu\textsuperscript{\rm 1,\rm 2},
Nan Min\textsuperscript{\rm 1},
Yuhang Guo\textsuperscript{\rm 1},
Tong Wei\textsuperscript{\rm 1,\rm 2\footnote{Corresponding author.}}
}
\affiliations {
\textsuperscript{\rm 1}School of Computer Science and Engineering, Southeast University, Nanjing, China\\ \textsuperscript{\rm 2}Key Laboratory of Computer Network and Information Integration (Southeast University), \\Ministry of Education, China\\     \{xuzhiy, mn614, gyh\_seu, weit\}@seu.edu.cn
}

\usepackage{bibentry}

\begin{document}

\maketitle
\begin{abstract}
3D Gaussian Splatting-based geometry reconstruction is regarded as an excellent paradigm due to its favorable trade-off between speed and reconstruction quality. However, such 3D Gaussian-based reconstruction pipelines often face challenges when reconstructing semi-transparent surfaces, hindering their broader application in real-world scenes. The primary reason is the assumption in mainstream methods that each pixel corresponds to one specific depth—an assumption that fails under semi-transparent conditions where multiple surfaces are visible, leading to depth ambiguity and ineffective recovery of geometric structures. To address these challenges, we propose TSPE-GS (Transparent Surface Probabilistic Extraction for Gaussian Splatting), a novel probabilistic depth extraction approach that uniformly samples transmittance to model the multi-modal distribution of opacity and depth per pixel, replacing the previous single-peak distribution that caused depth confusion across surfaces. By progressively fusing truncated signed distance functions, TSPE-GS separately reconstructs distinct external and internal surfaces in a unified framework. Our method can be easily generalized to other Gaussian-based reconstruction pipelines, effectively extracting semi-transparent surfaces without requiring additional training overhead. Extensive experiments on both public and self-collected semi-transparent datasets, as well as opaque object datasets, demonstrate that TSPE-GS significantly enhances reconstruction accuracy for semi-transparent surfaces while maintaining reconstruction quality in opaque scenes. The implementation code is available at \url{https://github.com/nortonii/TSPE-GS}.
\end{abstract}

\section{Introduction}

Geometric reconstruction represent fundamental challenges in graphics, essential for advancing applications such as augmented reality (AR), virtual reality (VR), autonomous driving \cite{zhou2024drivinggaussian,bruno20103d}. Recent research has seen the rise of Neural Radiance Fields (NeRF) as a leading approach for achieving precise geometric reconstruction \cite{mildenhall2021nerf}. Despite their effectiveness, NeRF models rely on volume rendering techniques that necessitate dense point sampling along rays, resulting in significant computational overhead and limiting real-time applications.

Recently, 3D Gaussian Splatting (3DGS) \cite{kerbl20233d} has emerged as a promising method for surface reconstruction, representing complex scenes as a set of 3D Gaussians. This technique has rapidly advanced to include surface reconstruction \cite{guedon2024sugar,huang20242d,yu2024gaussian,chen2024pgsr}. Notably, the SuGaR method \cite{guedon2024sugar} aligns 3D Gaussians with surfaces and employs Poisson surface reconstruction \cite{kazhdan2013screened} to extract meshes from depth maps. However, Poisson reconstruction has significant limitations, including the neglect of opacity and scale of Gaussian primitives, as well as unreliable depth maps that hinder real-time processing. To overcome these challenges, truncated signed distance functions (TSDF) \cite{curless1996volumetric} have been utilized for mesh reconstruction, extracting depth information by assessing the visibility of different Gaussian primitives. This approach encodes the distance to the nearest surface at localized points, allowing for integration of depth data from multiple viewpoints and facilitating continuous, smooth surface representations.

Despite the advancements in surface reconstruction using 3DGS, Reconstructing semi-transparent scenes remains challenging, both for the semi-transparent surfaces and the occluded objects behind them \cite{li2025tsgs}. These limitations are crucial for applications like robotic laboratory systems that require millimeter-precise manipulation of glassware, such as beakers and test tubes or tasks involving the retrieval of items from semi-transparent containers, such as food wrapped in cling film or objects inside plastic bags \cite{sajjan2020clear}. The first challenge in reconstructing semi-transparent surfaces arises from the common assumption in most existing depth extraction methods that there is only a single visible surface per ray. Consequently, these methods approximate the surface depth either as the weighted average (expected depth) \cite{chen2024pgsr} or a representative value (such as the median depth) \cite{zhang2024rade} of the visible Gaussian primitives’ depths. However, when multiple surfaces are visible simultaneously, the weighted average tends to fall between the actual surfaces, causing ambiguity. The second challenge arises from surface extraction methods utilizing TSDF depth fusion. If an object is sealed inside a plastic bag, even when employing current semi-transparent surface reconstruction techniques to extract the outermost depth from all viewpoints for TSDF fusion, the internal object remains unreconstructable, despite being clearly visible, as it does not exhibit depth from any viewpoint.

\begin{figure}[!ht]
    \centering
    \includegraphics[width=1.0\linewidth]{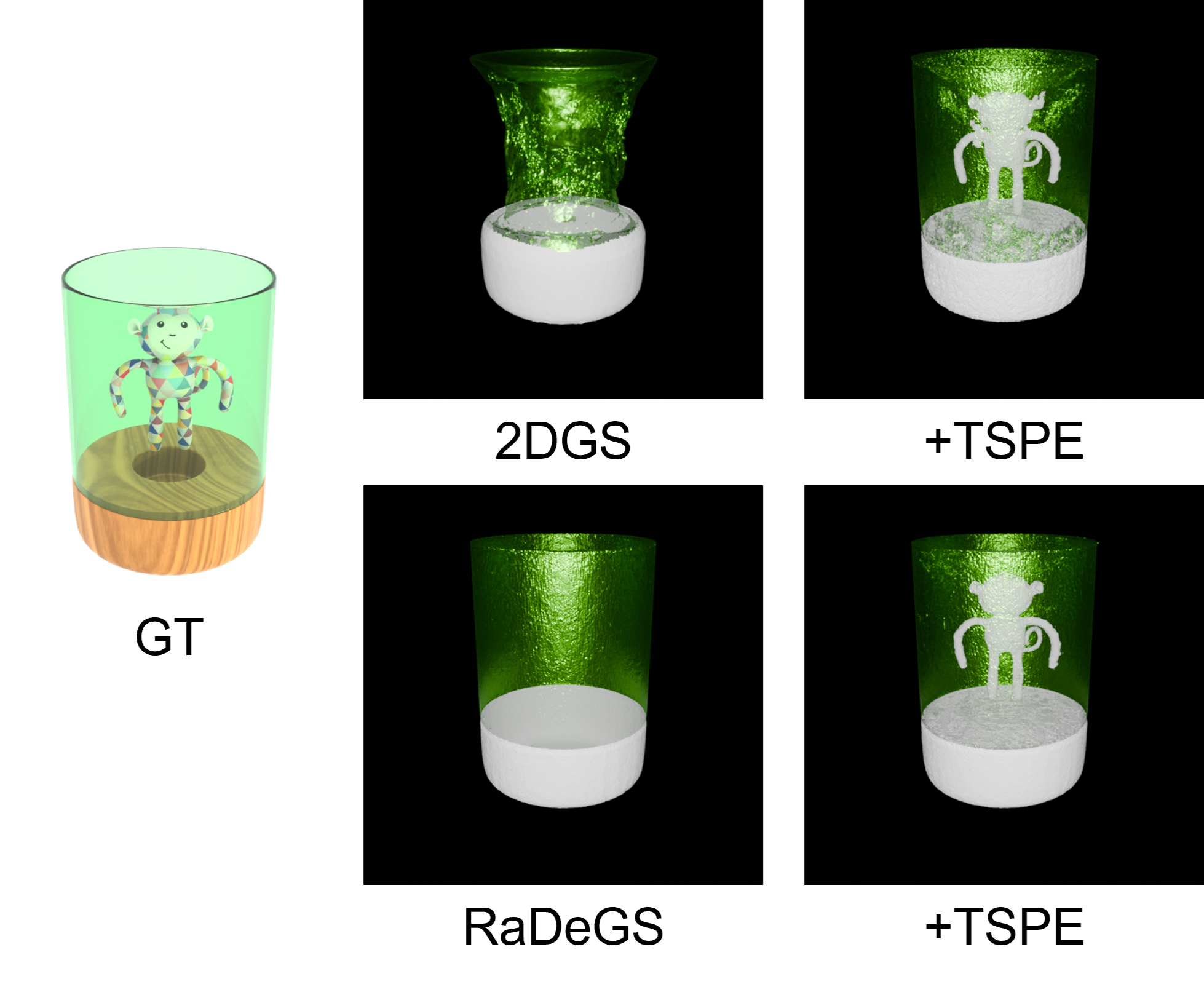}
    \caption{Visual comparison of reconstruction on a semi-transparent object in $\alpha$Surf dataset with and without TSPE.}
    \label{fig:firstvis}
\end{figure}

However, a simple yet effective approach exists that can simultaneously address the aforementioned two challenges. We argue that 3DGS can represent semi-transparent and occluded surfaces accurately, but current depth extraction methods underutilize this capability. The limitation lies primarily in the depth extraction methods, which have yet to fully exploit this potential. Specifically, the geometric expressiveness of 3DGS can be described as a mapping between visibility and depth values observed from multiple viewpoints. Existing depth extraction techniques have typically considered only the integral value (expected depth) or the median value of this mapping, overlooking the richer information contained within the full distribution. To address this, we propose an efficient method dubbed Transparent Surface Probabilistic Extraction for Gaussian Splatting (TSPE-GS). TSPE-GS captures the full visibility-depth mapping and applies a tailored analysis and processing strategy to simultaneously identify multiple visible surface depths. By leveraging tailored TSDF fusion across these depths, TSPE-GS enables concurrent and accurate reconstruction of both semi-transparent surfaces and occluded internal objects, as illustrated in Figure \ref{fig:firstvis}. We evaluate TSPE-GS on the BMVS, DTU, and $\alpha$Surf datasets \cite{yao2020blendedmvs,aanaes2016large,wu2025alphasurf}, along with our custom dataset designed specifically to include objects occluded by semi-transparent surfaces. Results show that TSPE-GS maintains reconstruction accuracy on general scenes (BMVS and DTU) without any degradation, while significantly improving reconstruction quality in semi-transparent surface scenarios ($\alpha$Surf and our dataset). Our method is simple and efficient, imposing negligible computational overhead relative to the highly optimized 3DGS rendering process. In summary, our contributions are as follows:
\begin{itemize}
\item We are the first to identify the challenge of robustly extracting objects occluded by semi-transparent surfaces within the 3DGS depth estimation reconstruction framework.

\item We propose a method to address this challenge by introducing a probabilistic depth extraction technique combined with a progressive TSDF fusion strategy that effectively reconstructs multi-layer surfaces.

\item Extensive experiments show that our method significantly outperforms state-of-the-art approaches in accuracy and completeness on both semi-transparent and opaque datasets, demonstrating its effectiveness and robustness.

\end{itemize}
\section{Related Work}
\textbf{Multi-view Stereo Surface Reconstruction} Multi-view stereo (MVS) depth estimation is a classical problem in computer vision and graphics~\cite{furukawa2009accurate, barnes2009patchmatch, bleyer2011patchmatch, schonberger2016structure,goesele2007multi,hirschmuller2007stereo}. Traditional methods extract dense point clouds from multi-view images through patch-based matching~\cite{barnes2009patchmatch} and reconstruct surfaces via triangulation~\cite{book}. While achieving high geometric accuracy, they often fail in weakly textured regions. Recent end-to-end neural networks~\cite{yao2018mvsnet, kendall2017end, luo2016efficient, gu2020cascade, ding2022transmvsnet} improve MVS by learning feature representations, but require costly volumetric representations and training with ground truth depth, limiting practical deployment in applications like robotics, AR/VR, and large-scale 3D reconstruction.

\textbf{Neural Surface Reconstruction} Unlike implicit methods requiring depth supervision, NeRF~\cite{mildenhall2021nerf} models scenes via MLPs mapping 5D coordinates to color and density, enabling photorealistic novel view synthesis. Extensions integrate signed distance functions (SDFs)~\cite{hart1989ray} for mesh extraction via marching cubes~\cite{li2023neuralangelo, yu2021plenoctrees, wang2021neus, wang2022neuris, yariv2021volume}, and NeuS~\cite{wang2021neus} bridged volume rendering with SDFs for unbiased surfaces. Neuralangelo~\cite{li2023neuralangelo} further improved large-scale reconstruction using multiresolution hash grids. To represent open surfaces, unsigned distance fields (UDFs) were proposed~\cite{liu2023neudf, long2023neuraludf}. Other works incorporate multi-view stereo principles by adding photometric consistency losses to enhance geometry~\cite{ding2022transmvsnet, fu2022geo}, but NeRF-based methods remain computationally expensive, limiting real-world use.

\textbf{Gaussian-Splatting based Surface Reconstruction} 3DGS represents scenes with Gaussian primitives, but extracting geometry is challenging due to its discrete and unstructured nature~\cite{kerbl20233d}. SuGaR~\cite{guedon2024sugar} first extracted meshes by encouraging Gaussian fitting and using Poisson reconstruction, but achieved limited accuracy lacking geometric regularization. The idea of approximating 3D Gaussians as planes (shortest axis as normal) was extended in 2D Gaussian Splatting (2DGS)~\cite{huang20242d} and PGSR~\cite{chen2024pgsr}—the former compressing 3DGS into 2D planes, the latter emphasizing cross-view geometric consistency, though compressing reduces rendering quality. GOF~\cite{yu2024gaussian} defines a Gaussian opacity field using maximum Gaussian-ray intersection depth, but pixel-wise depth calculation adds heavy computation. RaDeGS~\cite{zhang2024rade} improved on this by proving the level set normal is planar, reducing training time and improving geometry, yet introduced new approximations and bias. Inspired by these, we propose a geometry extraction method without approximations and additionally incorporate multi-view geometric consistency to optimize depth and normals, achieving high-precision reconstruction with comparable computational resources.
\section{Preliminaries}
\textbf{Render process:} 3DGS fits a scene using a set of Gaussian primitives $\mathcal{G}=\{g_i|i=1, .., N\}$. Each Gaussian primitive $\mathcal{G}_i$ is parameterized by a full 3D covariance matrix $\Sigma_i \in \mathbb{R}^{3 \times 3}$, which is centered at the world space point $\mu_i \in \mathbb{R}^3$:

\begin{equation}
  g_i(x) = e^{-\frac{1}{2}(x - \mu_i)^T(\Sigma_i)^{-1}(x - \mu_i)}
\end{equation}

Next, spherical harmonics are assigned to each Gaussian to model view-dependent colors and employ $\alpha$-blending to integrate colors in sequence:
\begin{equation}
  c = \sum_{i=1}^N \alpha_i \prod_{j=1}^{i-1}(1 - \alpha_j) c_i
\end{equation}
where $\alpha_i = g_i \cdot o_i$ is parameterized by the covariance matrix multiplied by a learnable opacity $o_i$ for each Gaussian.

\textbf{Depth extraction:} In depth extraction, based on different 3DGS methods, depth extraction can be classified into two forms: Expected depth and Median depth.
\begin{itemize}
\item Expected depth $D^{\text{exp}}_p$ is defined as:
   \begin{equation}
   D^{\text{exp}}_p = \sum_{g \in \mathcal{G}_{p}} T_{g-1} \alpha_g d_{p,g}.       
   \end{equation}
\item Median depth $D^{\text{med}}_p$ is defined as:
   \begin{equation}
   D^{\text{med}}_p = \max_{g \in \{ g \, | \, T_{p,g} < 0.5 \}} d_{p,g}.   
   \end{equation}
\end{itemize}
where \( \mathcal{G}_{p} \) represents the set of Gaussian primitives associated with pixel \( p \), \( T_g = \prod_{j=1}^{g-1}(1 - \alpha_j) \) denotes the accumulated transmittance for Gaussian \( g \), and \( d_{p,g} \) is the depth of Gaussian primitive \( g \) at pixel \( p \). 
\section{The Proposed Method}
We next detail the core components of TSPE-GS, starting with probabilistic depth modeling.
\subsection{Probabilistic depth modeling}

\begin{figure}[ht]
\centering
\includegraphics[width=0.45\textwidth]{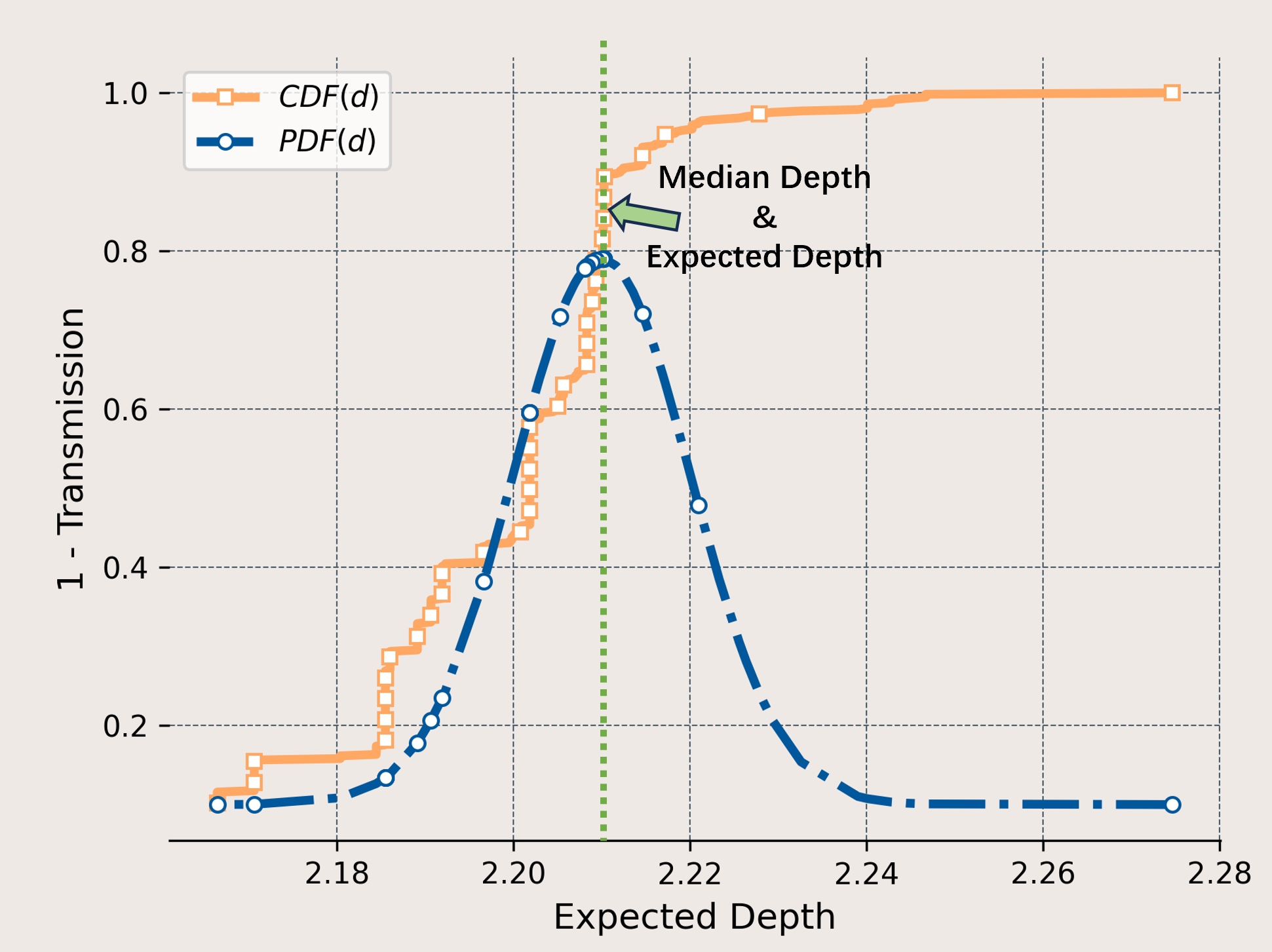} 
\caption{Depth PDF and CDF for opaque surfaces.}
\label{opad2t}
\end{figure}

First, based on the previous definitions, we can derive the following formula that represent a  relationship between \( T_p \) and \( d_p \):
\begin{equation}
 d_p \to T_p:  \quad T_p(d_p) = \min_{g \in \{ g \, | \, d_{p,g} > d_p \}} T_{p,g} 
\end{equation}
Next, we can extend this mapping to a continuous form. For a specified pixel \( p \), we assign an opacity \( \alpha_d \) to each infinitesimal depth element within the depth range \( d \). Under the constraints of 
\begin{equation}
\int_{T_{g-1}}^{T_{g}} d_T \mathrm{d}T = \Delta T_g d_g \label{cons1}   
\end{equation}
and 
\begin{equation}
\alpha_g = 1 - e^{\int_{d \in g} \ln(1 - \alpha_d) \, \mathrm{d}d}, \label{cons2}
\end{equation}
we can express the transmittance \( T_d \) corresponding to each depth \( d \) for pixel \( p \) as: $T_d = \prod_0^d (1 - \alpha_d)$. It is straightforward to verify that, under this definition, the following properties hold: \( T_{p,i} \geq T_{p,j} \) only if \( d_i \geq d_j \), the transmittance \( T \) approaches 0 as depth tends to infinity, and approaches 1 as depth approaches zero. These conditions naturally suggest modeling the relationship between \( T \) and \( d \) using a cumulative distribution function (CDF), which can be expressed as: \begin{equation}
    P(d_{p,g} < d) = 1 - T_d
\end{equation}when defining this probabilistic framework, we also obtain a definition for the probability density function (PDF):
\begin{align}
p(d) & = -\frac{\mathrm{d} T_d}{\mathrm d d}  = -\frac{\mathrm de^{\int_0^d\mathrm{ln}(1-\alpha_d)\mathrm{d}d }}{\mathrm d d}\nonumber\\
&=-\mathrm{ln}(1-\alpha_d)T\approx\alpha_d T_d
\end{align}
Noting that each infinitesimal \(\alpha_d\) is sufficiently small, so \(\lim_{x \to 0} \ln(1 - x) \to -x\). This approximation shows that \(p(d)\) matches precisely the weighting used in the computation of expected depth, ensuring consistency between the probabilistic framework and depth expectation. According to experiments in Appendix A, \( p(d) \) exhibits a unimodal distribution very close to a Gaussian when no transparent surfaces are present, while the CDF displays a classic S-shaped curve, as illustrated in Figure \ref{opad2t}.

The median depth clearly corresponds to the depth at which the CDF equals 0.5. However, for expected depth, some transformations are required: Referring to the above definitions, the relationship can be expressed as follows:
\begin{equation}
T_g = T_{g-1} \cdot (1 - \alpha_g)    
\end{equation}
Consequently, \(T_{g-1} \alpha_g\) can be reformulated as:
\begin{equation}
T_{g-1} \alpha_g =  T_{g-1}-T_g  
\end{equation}
This leads to a reformulation of expected depth. By substituting the constraint given in Eq. \ref{cons1}, we obtain:
\begin{equation}
    D^{\text{exp}}_p = \sum_{g \in \mathcal{G}_{p}}  \Delta T_g d_{p,g}=\sum_{g \in \mathcal{G}_{p}}\int_{T_{g-1}}^{T_g} d_T \mathrm{d}T =\int_0^{1}d_T\mathrm{d}T
\end{equation}
Subsequently, we perform a change of variables to obtain the following form:
\begin{equation}
D^{\text{exp}}_p = \int_0^{d_{\max}}d\frac{\mathrm{d}T_d}{\mathrm{d}d}\mathrm{d}d=\int_0^{d_{\max}}dp(d)\mathrm{d}d=E(d)
\end{equation}
Thus, the expected depth is the expected value of the previously defined depth distribution. Since it is a unimodal distribution similar to a Gaussian distribution, the point where the CDF equals \(0.5\) and the expected value both clearly lie at the peak of the unimodal distribution, which corresponds to the mean of the Gaussian distribution, as shown in Figure \ref{opad2t}. Currently, many 3D surface reconstruction methods are based on these two types of depth, considering that both depths closely align with the target surface height. Therefore, it can be inferred that in non-transparent scenarios, the position of peak points represents the location of surfaces.

\begin{figure}[ht]
    \centering
    \includegraphics[width=0.45\textwidth]{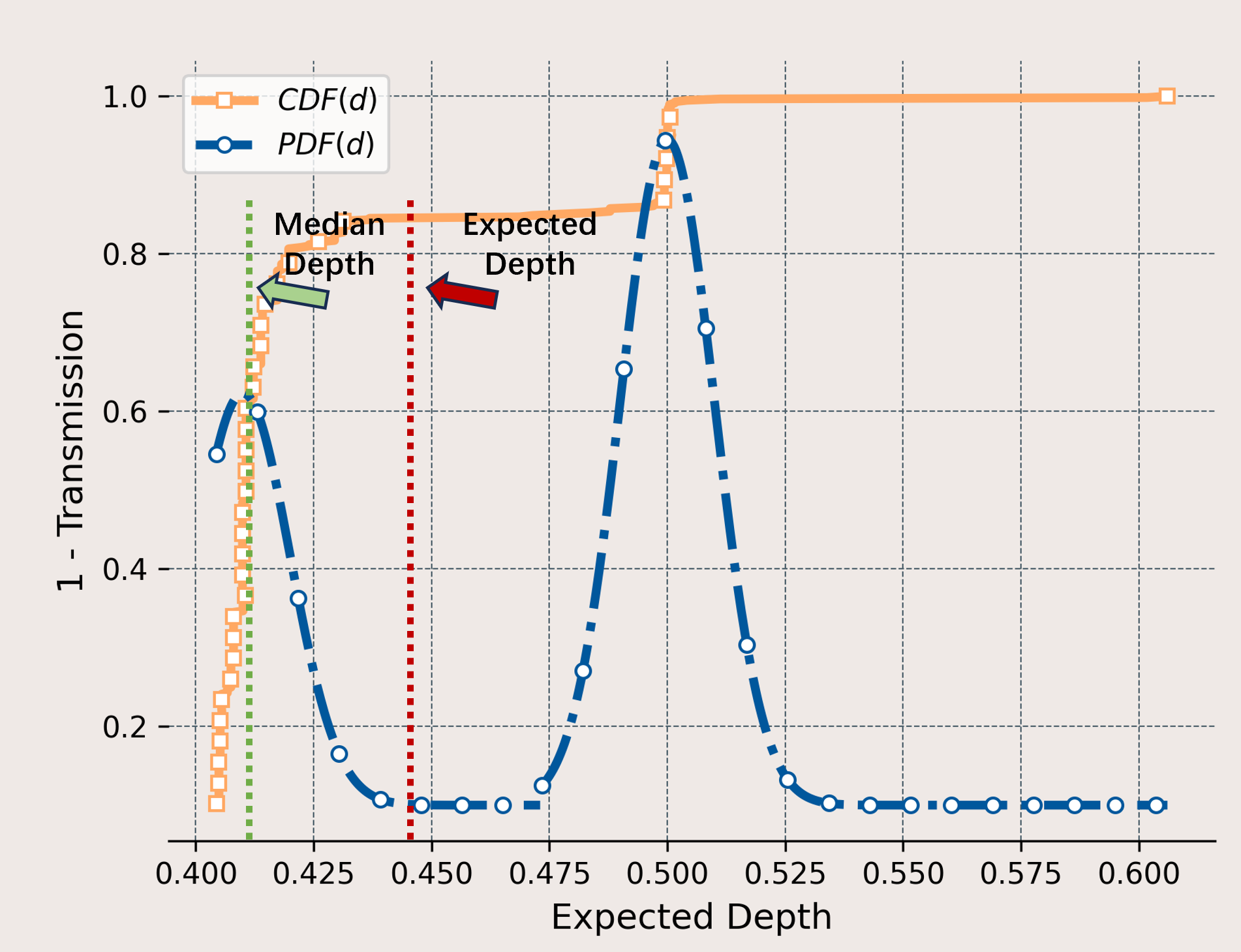}
    \caption{Depth PDF and CDF for semi-transparent surfaces.}
    \label{transd2t}
\end{figure}

This corresponds to the situation in the absence of semi-transparent surfaces. Surprisingly, even in the presence of multiple visible surfaces, as shown in Figure \ref{transd2t}, the depth distribution exhibits multiple peaks resembling a mixture of Gaussian distributions. These peaks reliably correspond to visible surfaces, as consistently demonstrated by empirical experiments. Further details of these experiments can be found in Appendix B.

Based on these observations, we propose a unified hypothesis that accounts for surface depth in both semi-transparent and non-transparent scenarios:

\paragraph{Unified surface depth hypothesis}  
The depth probability distribution for a given pixel can be modeled as a (uni- or multi-)modal distribution whose peaks correspond directly to the physical surface depths along the viewing ray. Formally, whether the distribution is unimodal (absence of semi-transparent surfaces) or multimodal (presence of semi-transparent surfaces), the set of surface depth locations \(\{\mu_k\}\) are given by the set of local maxima of the distribution:
\begin{equation}
\{\mu_k\} = \{ d \mid p'(d) = 0, \quad p''(d) < 0 \}
\end{equation}
where each \(\mu_k\) corresponds to one visible surface depth. This hypothesis  generalizes the classical unimodal assumption to handle complex scenes involving semi-transparent layers.

However, in the presence of semi-transparent surfaces, both the expected depth and the median depth fail to accurately represent the depths of the multiple visible layers. This is mainly because when a pixel corresponds to multiple surfaces, these single-value statistics can only reflect one layer at most. Consider a bimodal depth distribution modeled as a mixture of Gaussians:$p(d) = w_1 \cdot \mathcal{N}(d; \mu_1, \sigma_1^2) + w_2 \cdot \mathcal{N}(d; \mu_2, \sigma_2^2)$, where \(\mathcal{N}(d; \mu, \sigma^2)\) is a Gaussian with mean \(\mu\) and variance \(\sigma^2\), and \(w_1 + w_2 = 1\) are the mixing weights.

Assuming that the two components are well separated with large variances, the peaks naturally correspond to \(\mu_1\) and \(\mu_2\), representing the visible surface depths. The expected depth, calculated as \(D_{exp} = w_1 \mu_1 + w_2 \mu_2\), lies between these peaks, losing direct physical interpretability. Similarly, the median depth also falls between the two peaks, complicating the interpretation of depth for semi-transparent surfaces (see Figure \ref{transd2t}).

To overcome these challenges, we propose TSPE-GS, a novel method based on the key observation that the peaks of the depth probability distribution correspond directly to physical surface locations. Our approach for reconstructing scenes with semi-transparent surfaces involves three key steps: 1) obtain the probability distribution; 2) calculate the locations of the peaks; and 3) fuse peak depths.

\subsubsection{Obtaining the Probability Distribution}

To compute the probability distribution \(P(d)\), our objective is to establish the relationship between depth \(d\) and  quantity \(1 - T\), while ensuring that the two constraints are satisfied.
\begin{figure*}[ht]
    \centering
    \includegraphics[width=0.9\linewidth]{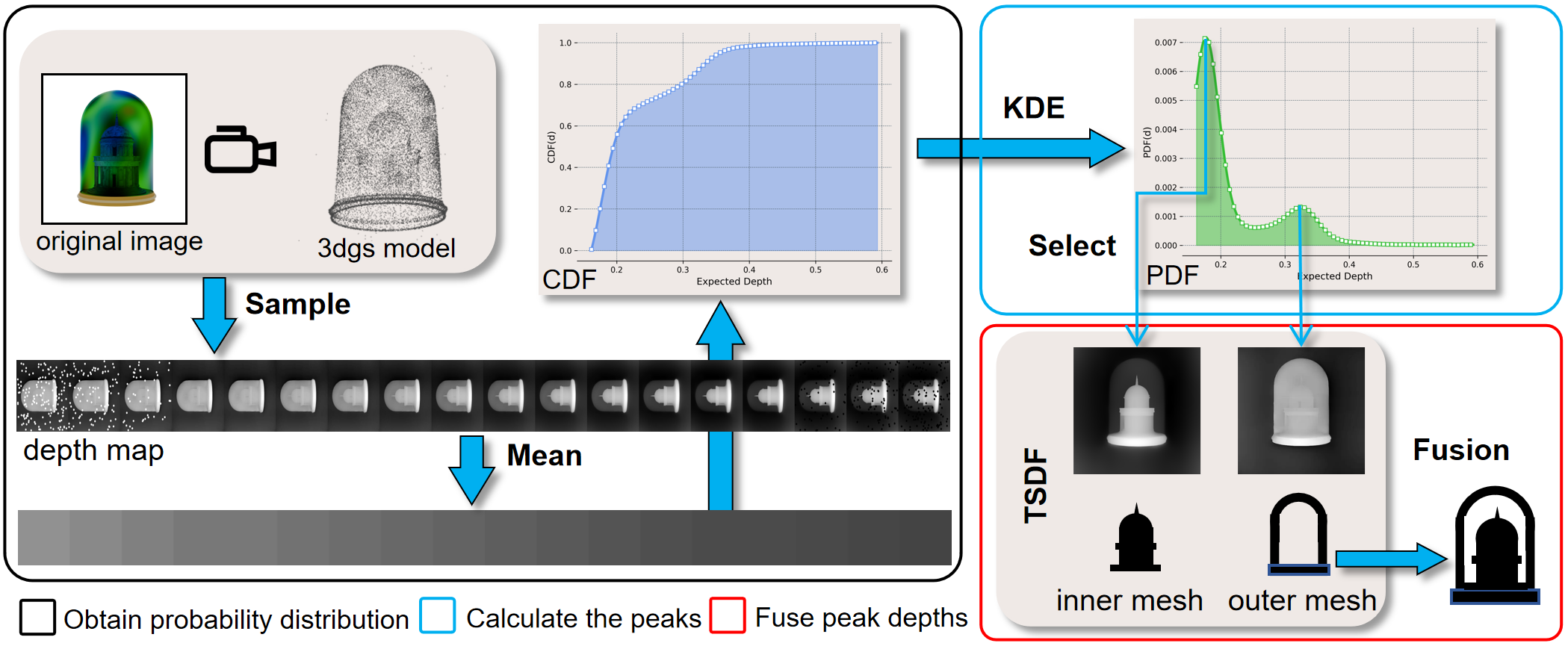}
    \caption{Workflow of the TSPE-GS pipeline for multi-layer depth estimation and reconstruction.}
    \label{pipline}
\end{figure*}
We can define $\alpha_d = 1 - e^{\frac{\ln(1 - \alpha_g)}{d_g - d_{g-1}} (d - d_{g-1})}$, which holds for any \(d\) such that \(d_{g-1} < d < d_g\). This definition of \(\alpha_d\) simultaneously satisfies constraints in Eq. \ref{cons1} and Eq. \ref{cons2}. Because we need to sample many values of \(T\) with corresponding depths \(d\) and then use density estimation to calculate the PDF, high-precision depth values are not required. Therefore,  Instead of deriving \(d\) from \(\alpha_d\), we set \(d = d_g\) whenever \(T_g < T_d < T_{g-1}\). Experiments show that the PDF obtained by this simplification closely matches the strict PDF. More detailed experiments and derivations can be found in Appendix C.

Therefore, to obtain the required depths for different values of \(T'\), we only need to compute $D'_p = \max_{g \in \{ g \mid T_{p,g} < T' \}} d_{p,g}$, which represents the maximum depth \(d_{p,g}\) among all indices where the transmittance \(T_{p,g}\) is less than the threshold \(T'\). This method is easy to incorporate into existing 3DGS frameworks. Below is an example pseudocode illustrating how to compute \(D'_p\) for a given \(T'\):
\begin{algorithm}[!ht]
\caption{3D Gaussian Splatting Depth Extraction}
\label{alg:3dgs_color_depth_accum}
\textbf{Input}: For each pixel \(p\), lists of transmittance values \(T_{p,g}\), depths \(d_{p,g}\), colors \(c_g\), opacities \(\alpha_{p,g}\) for each Gaussian \(g\)\\
\textbf{Parameter}: Opacity threshold \(\tau\), transmittance threshold \(T'\)\\
\textbf{Output}: Color and depth \(C_p\), \(D'_p\)
\begin{algorithmic}[1]
\FOR{each pixel \(p\)}
  \STATE Initialize \(C_p \leftarrow 0\), \(\alpha_p \leftarrow 0\), \(D'_p \leftarrow 0\)
  \FOR{each Gaussian \(g\) intersecting pixel \(p\)}
    \STATE Compute transmittance \(T_{p,g} = \prod_{k=1}^{g-1} (1 - \alpha_{p,k})\)
    \STATE Compute opacity contribution \(\alpha_{p,g}\)
    \STATE Update opacity: \(\alpha_p \leftarrow \alpha_p + (1 - \alpha_p) \alpha_{p,g}\)
    \STATE Update color: \(C_p \leftarrow C_p + (1 - \alpha_p) \alpha_{p,g} c_g\)
    \IF{ \(T_{p,g} < T'\) }
      \STATE Assign depth: \(D'_p \leftarrow d_{p,g}\)
      \STATE \textbf{break}
    \ENDIF
  \ENDFOR
  \STATE Output \(C_p, D'_p\)
\ENDFOR
\end{algorithmic}
\end{algorithm}

In this work, the sampling of \(T'\) values is performed using uniform sampling, defined as
$T' \in \left\{ \frac{k}{N} \mid k = 1, 2, \ldots, N \right\}$. Thanks to the extremely fast rendering speed of 3DGS, the sampling process for all viewpoints can be completed within approximately $1 \sim 2$ minutes. Finally, The depth distribution \(p(d)\) is obtained by applying kernel density estimation to the sampled depth values.
\subsection{Calculating the Peak Locations}

Having obtained the probability distribution \(p(d)\) of depth \(d\) for each pixel across all viewpoints, surface depth can be estimated by detecting peaks in this distribution. However, this approach faces two main challenges: the presence of abnormal Gaussian primitives causing spurious multiple peaks, and the high computational cost of performing peak detection on every pixel at all viewpoints.

However, we found a useful property to address these challenges. By averaging depth values \(d\) across all pixels at each of the \(N\) transmittance thresholds \(T'\) for a viewpoint, we get a depth sequence of length \(N\) that exhibits similar multi-peak characteristics as the per-pixel distributions, conforming to the unified surface depth hypothesis (see Figure \ref{pipline}). To mathematically explain this, we categorize all visible surfaces into \(\mathcal{S}\) types, where each type corresponds to surfaces with distinct materials or occlusion relationships. For a surface type \(s \in \mathcal{S}\), the depth distributions at different pixels belonging to that surface form a family of functions \(p_p(d) = p(d + t)\), where \(t\) denotes the spatial shift due to different positions of the surface in the scene.

For any two different spatial shifts \(t_0\) and \(t_1\), corresponding to depth distributions \(p_{p,0}\) and \(p_{p,1}\), let their extremum points be \(d_0\) and \(d_1\). The \( \text{CDF}_{p,0}(d_0) \) and \( \text{CDF}_{p,1}(d_1) \) both equal the same transmittance threshold \(T'\).

Now, define an interpolated inverse cumulative distribution function (ICDF) as: $\mathrm{ICDF}(T) = a \cdot \mathrm{ICDF}_0(T) + (1 - a) \cdot \mathrm{ICDF}_1(T)$
, for \(a \in [0,1]\). The probability distribution corresponding to this ICDF satisfies $\mathrm{CDF}(d') = T'$, meaning that \(d'\) is also an extremum point of this combined distribution. The proof is as follows: from the definition of the inverse CDF, \( \text{ICDF}(T) \) is the quantile such that \( F(\text{ICDF}(T)) = T \). Differentiating this w.r.t. \( T \) gives:  
\begin{equation}
    p(d) = \frac{1}{\text{ICDF}}=\frac{1}{a \cdot \text{ICDF}_0' + (1 - a) \cdot \text{ICDF}_1'}
\end{equation}
To identify extrema of \( p(d) \), we examine where the derivative \( \frac{\mathrm{d}}{\mathrm{d}d} p(d) \) vanishes. Using the chain rule and noting \( d = \text{ICDF}(T) \), this corresponds to the condition:  
\begin{equation}
p(d)\frac{\mathrm{d}}{\mathrm{d}T} \left( a \cdot \text{ICDF}_0'(T) + (1 - a) \cdot \text{ICDF}_1'(T) \right) = 0.  
\end{equation}
The second derivative of the ICDF is related to the derivative of the PDF by: 
\begin{equation}
\text{ICDF}''(T) = - \frac{p'(\text{ICDF}(T)) \cdot \text{ICDF}'(T)}{p(\text{ICDF}(T))^2}.  
\end{equation}
Evaluating at \( T = T' \), and noting that \( p_{p,0}'(d_0) = 0 \) and \( p_{p,1}'(d_1) = 0 \), it follows that  
\begin{equation}
\frac{d}{dT} \left( a \cdot \text{ICDF}_0''(T) + (1 - a) \cdot \text{ICDF}_1''(T) \right) \bigg|_{T=T'} = 0.    
\end{equation}
Therefore, the derivative of the denominator of \( p(d) \) is zero at \( d' = \text{ICDF}(T') \), hence \( p(d) \) has a stationary point at \( d' \). This implies that for the same surface type, there exists a transmittance threshold \(T'\) such that the aggregated depth value \(d' = \mathrm{ICDF}(T')\), which corresponds to the mean over pixels, is an extremum point of the combined distribution. Moreover, the depth at each pixel, \(d_p = \mathrm{ICDF}_p(T')\), corresponds to the extremum point of its individual distribution. As a result, using the mean depth values computed from each depth map at a given transmittance \(T\) is sufficient to identify all surface depth. Following the same approach, we construct a PDF for the combined depth distribution.

After obtaining the new PDF, we use a local maxima detection algorithm to estimate a peak score for each sampled point and define peaks with scores above a certain threshold as target peaks. Selecting an appropriate threshold is crucial: too high a value may miss true visible surface depths, while too low a value can introduce floating artifacts by including spurious peaks. Thus, the threshold should balance capturing genuine surface depths and minimizing false positives for robust extraction. We further analyze this influence in the experiments section.

\subsection{Fuse Peak Depths}
After extracting peak depths from probabilistic distributions, we fuse the resulting multi-layer depth maps into a volumetric grid using TSDF. Each peak depth map is integrated into the TSDF volume according to camera poses, with TSDF values and weights updated via weighted averaging.
However, directly fusing all peak depth maps can cause problems such as surface roughness and holes due to interference between layers. This occurs because inner layer depths—corresponding to occluded surfaces—spatially overlap with outer layer depths covering the same regions. Since inner layer depths are generally less accurate, their interference degrades the quality of outer surface reconstruction. Further analysis are provided in Appendix D.
To mitigate this, we propose a progressive reconstruction strategy: first reconstruct the outermost surfaces and freeze their associated voxels in the TSDF volume; then reconstruct inner surfaces while preventing interference from the frozen outer layers. The pseudocode is detailed in Appendix E.
In summary, our method combines probabilistic depth modeling, multi-peak detection, and progressive TSDF fusion to reconstruct multi-layer surfaces in scenes with transparency and occlusion. Figure \ref{pipline} illustrates the overall workflow.

\begin{table}[!ht]
\centering
\footnotesize
\setlength{\tabcolsep}{2pt}
\begin{tabular}{lccccccccc}
\hline
\multicolumn{1}{c}{{\textbf{Method}}} & Cas & Tab & Cof & Kit & Bot & Mon & Dou & Vas & \textbf{Avg. $\downarrow$} \\
\hline
Mip360& 4.01 & 5.22 & 2.10 & 2.38 & 2.32 & 4.39 & 2.90 & 1.19 & 3.13 \\
NeuS & 5.09 & 1.19 & 1.07 & 0.39 & 2.27 & 5.92 & 0.87 & 1.95 & 2.34 \\
HFS & 5.09 & 0.85 & 2.84 & N/A & 1.49 & 3.22 & N/A & 8.68 & N/A \\
angelo& 2.07 & 0.48 & 0.80 & 0.58 & {0.40} & 2.46 & {0.51} & 1.70 & 1.12 \\
NeRRF & 1.27 & 0.57 & 0.82 & 3.09 & 3.72 & 2.90 & 0.83 & 3.49 & 2.11 \\
\hline
PGSR & 1.44 & 0.60 & 3.60 & 2.12 & 4.83 & 12.20 & 2.24 & 0.86 & 3.48 \\
GOF & 0.64 & 0.85 & 1.08 & 0.96 & 0.84 & 2.80 & 1.19 & 0.71 & 1.13 \\
GSF & 1.52 & 1.74 & 0.61 & 0.82 & 0.62 & 3.00 & 0.87 & 1.63 & 1.39 \\
\hline
2DGS & 1.21 & 0.88 & 1.14 & 1.00 & 2.05 & 7.70 & 1.38 & 1.13 & 2.06 \\
+TSPE & 0.72 & 1.45 & 0.99 & 0.70 & 1.45 & 1.47 & 1.12 & 0.72 & \bf{1.08} \\
RaDe & 0.49 & 0.57 & 0.98 & 0.44 & 0.80 & 3.00 & 0.63 & 0.85 & 0.97 \\
+TSPE & 0.48 & 0.51 & 1.04 & 0.51 & 0.76 & 1.66 & 0.56 & 0.76 & \bf{0.82} \\

\hline
\end{tabular}
\caption{Quantitative comparison on the $\alpha$Surf dataset \cite{wu2025alphasurf} for translucent object reconstruction evaluated by Chamfer distance ($\times10^{-2}$). Note that HFS \cite{wang2022hf} fails to learn any surface on ``kitchen table" and ``double table" scenes.}
\label{table:merged_cleaned_with_newmethod}
\end{table}
\begin{table}[!ht]
\centering
\footnotesize
\setlength{\tabcolsep}{2pt}
\begin{tabular}{lccccccccc}
\hline
\textbf{Method} & BLO & BOS & CAP & COM & FTR & GUM & HFD & SBL & \textbf{Avg.$\downarrow$} \\
\hline

2DGS        & 2.35 & 6.23 & 1.03 & 3.68 & 2.28 & 1.53 & 3.42 & 1.66 & 2.65 \\
PGSR        & 2.41 & 0.43 & 0.71 & 3.01 & 1.28 & 1.37 & 2.76 & 0.70 & 1.58 \\
GOF         & 2.40 & 1.20 & 0.50 & 1.99 & 1.22 & 1.56 & 3.31 & 1.10 & 1.66 \\
TSGS        & 1.99 & 0.64 & 0.50 & 2.80 & 2.14 & 1.61 & 3.57 & 1.41 & 1.83 \\
\hline
RaDe       & 2.43 & 0.49 & 0.48 & 1.98 & 1.23 & 1.07 & 4.79 & 1.19 & 1.71 \\
+TSPE      & 0.96 & 0.59 & 0.69 & 1.51 & 0.74 & 0.58 & 2.51 & 0.48 & \bf{0.88} \\
\hline
\end{tabular}
\caption{Quantitative results for various methods evaluated on the bottleship dataset. The Avg. column shows the mean number of valid entries per row.}
\label{table:method-comparison-3letter-codes}
\end{table}
\section{Experiments}
\textbf{Datasets.} To validate the effectiveness of our method in reconstructing semi-transparent surfaces, we conducted experiments on two semi-transparent datasets: the publicly available 
$\alpha$surf \cite{wu2025alphasurf} dataset and our self-captured semi-transparent dataset — which contains objects occluded by semi-transparent surfaces and is casually named the ``bottleship'' dataset. In addition, we also evaluated the reconstruction performance of our method on opaque surface datasets, primarily focusing on two classic geometric reconstruction benchmarks: DTU \cite{aanaes2016large} and BlendedMVS \cite{yao2020blendedmvs}.

\noindent
\textbf{Baselines.} To position our method within the field of semi-transparent surface reconstruction, we benchmarked our proposed approach against state-of-the-art (SOTA) methods, including 3D Gaussian-based reconstruction pipelines and NeRF-based geometry reconstruction pipelines.

\noindent
\textbf{Metrics.} Our method aims to evaluate geometry reconstruction quality, with Chamfer Distance (CD) serving as the primary evaluation metric.

\noindent
\textbf{Implementation.} During the training phase, we adopted the latest open-source implementations of the baselines and strictly adhered to their hyperparameter settings, including training epochs and optimizers. While in the geometry extraction stage, to ensure consistency, we uniformly employed the TSDF method for geometric reconstruction across all methods.

\begin{figure}[!ht]
    \centering
    \includegraphics[width=0.95\linewidth]{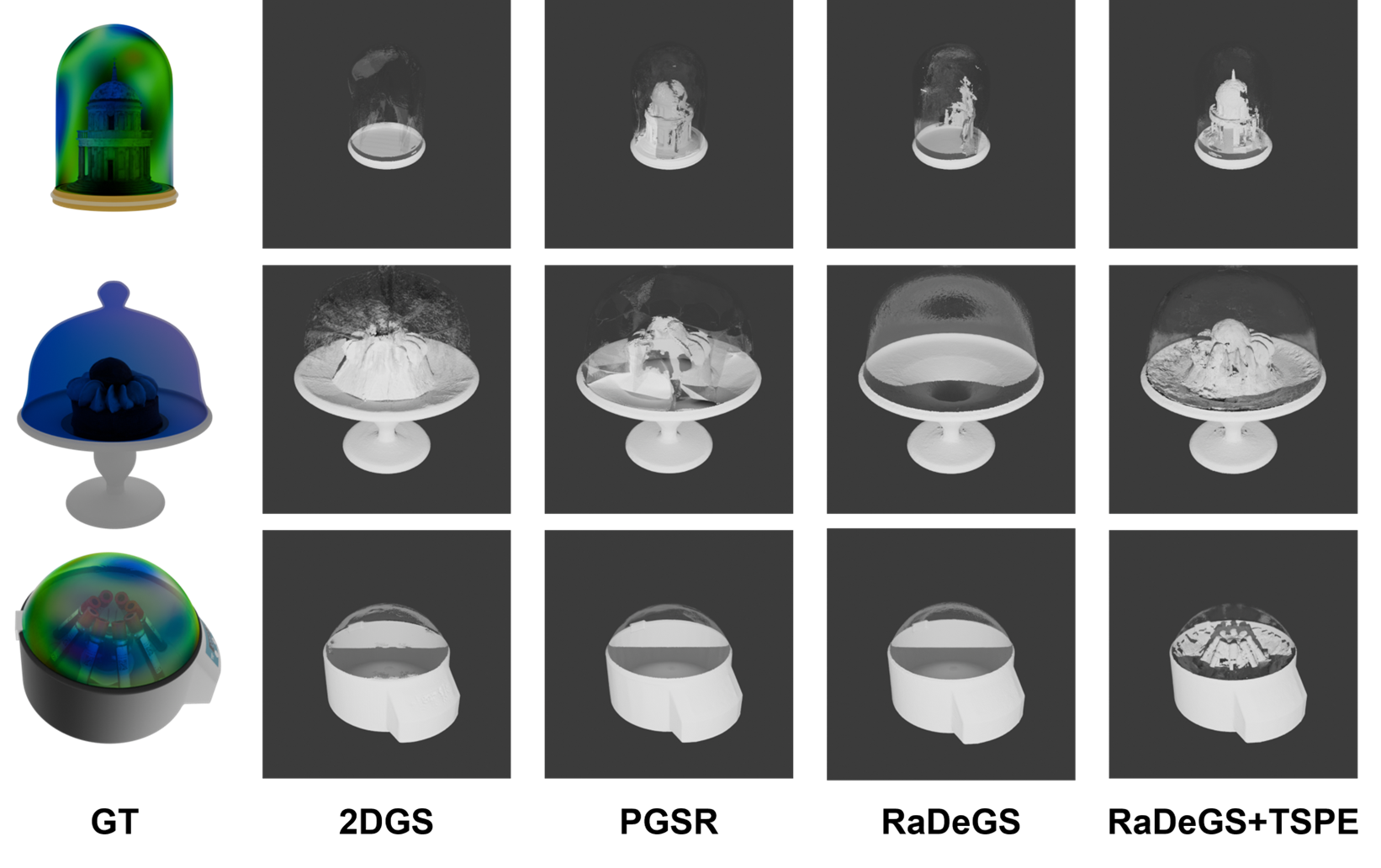}
    \caption{We visually compare our method with other Gaussian-based geometry reconstruction pipelines to demonstrate the effectiveness.}
    \label{vis}
\end{figure}
\begin{figure}[ht]
    \centering
    \includegraphics[width=0.8\linewidth]{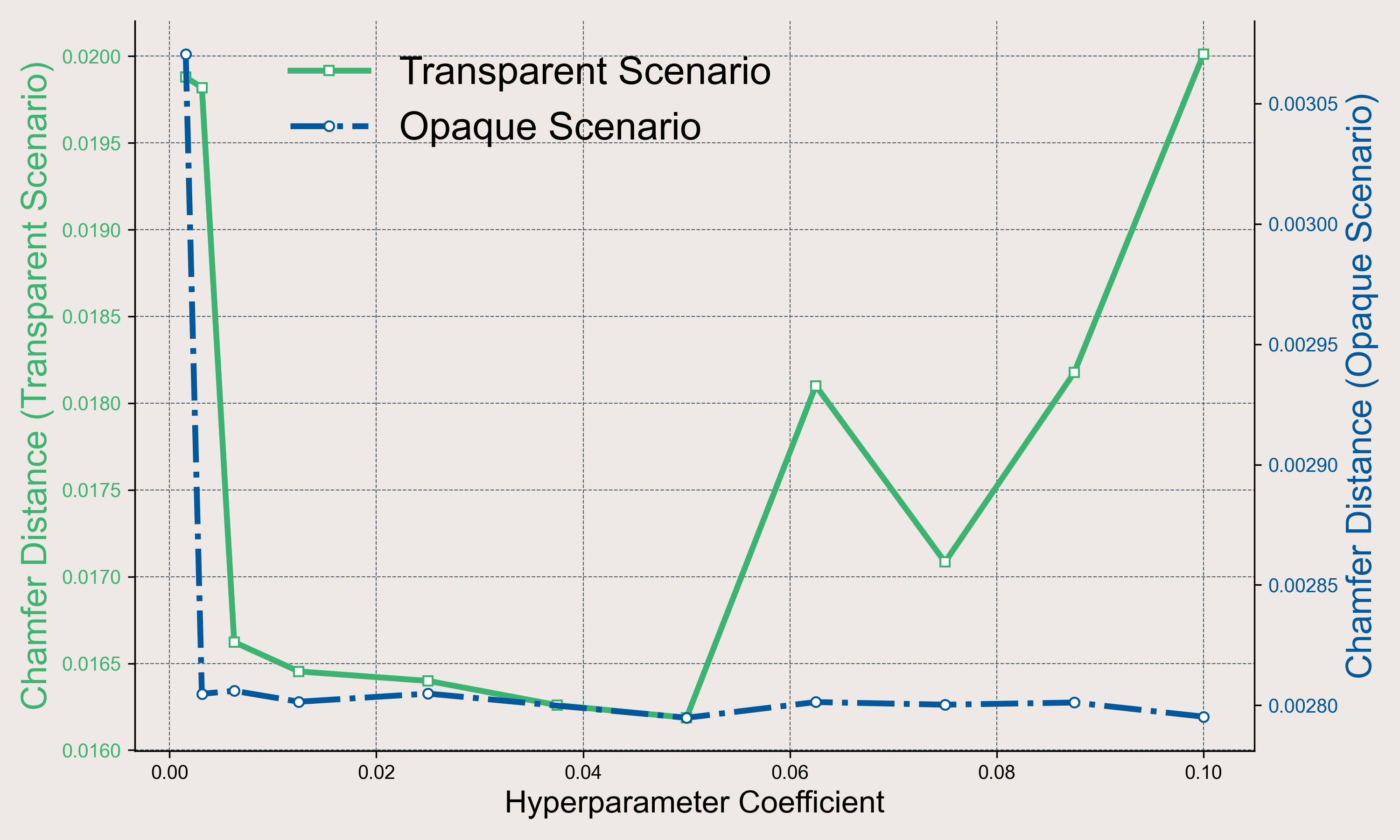}
    \caption{Impact of peak threshold on chamfer distance in semi-transparent and opaque scenarios.}
    \label{fig:hyper}
\end{figure}

\subsection{Result Comparison} Table 1 and Table 2 present the reconstruction quality of various methods on semi-transparent datasets using NeRF-based and 3D Gaussian-based pipelines. Our approach demonstrates distinct advantages within the 3D Gaussian pipeline framework, achieving state-of-the-art results with RaDeGS as the backbone across both datasets. Additionally, our proposed method demonstrates competitive performance within the NeRF-based geometry reconstruction pipeline. Furthermore, to validate the generalization capability of our approach, Tables 3 and 4 present experimental results on opaque datasets using several backbone architectures. These results empirically demonstrate the effectiveness of our method in reconstructing opaque surfaces. However, we observe that not all 3D Gaussian-based methods are equally compatible with TSPE; a fundamental requirement is that the underlying method must effectively represent the scene. For example, 2DGS performs poorly on our bottleship dataset, limiting TSPE’s applicability. Detailed analyses are provided in Appendix F.

\subsection{Visual Comparison} In Figure 6, we compare our method with classical Gaussian-based geometric reconstruction methods. It can be observed that these SOTA geometric reconstruction methods tend to represent either a specific plane—either the outer or inner surface—failing to reconstruct both surfaces simultaneously. However, with the integration of our extraction module, we effectively extract both outer and inner surfaces.
\subsection{Hyperparameter Analysis}  
We investigate the sensitivity of the peak threshold coefficient, a key hyperparameter in our probabilistic depth estimation process. As shown in Figure \ref{fig:hyper}, both semi-transparent and opaque scenarios share a common range of values that achieve near-optimal Chamfer Distance results. This overlap indicates that our method is robust to the choice of this hyperparameter across diverse scene types.
\begin{figure}[!ht]
  \centering
  
  \begin{minipage}[t]{0.48\columnwidth}
  \vspace{0pt}
    \centering
    \footnotesize
    \setlength{\tabcolsep}{2pt}
    \begin{tabular}{l|cc|cc}
    \hline
    \textbf{Scene} & \textbf{RaDe} & \textbf{\tiny{+TSPE}} & \textbf{2DGS} & \textbf{\tiny{+TSPE}} \\
    \hline
    Bea  & 0.50 & 0.56 & 1.08 & 0.66 \\
    Clo  & 0.87 & 0.87 & 0.35 & 0.26 \\
    Dog  & 1.07 & 1.06 & 0.64 & 0.33 \\
    Dur  & 0.49 & 0.49 & 5.72 & 2.91 \\
    Jad  & 0.14 & 0.14 & 0.29 & 0.14 \\
    Man  & 2.89 & 2.91 & 0.75 & 1.12 \\
    Scu  & 0.20 & 0.20 & 0.73 & 0.80 \\
    Sto  & 0.28 & 0.28 & 2.32 & 0.93 \\
    \hline
    Avg$\downarrow$ & 0.81 & \textbf{0.81} & 1.48 & \textbf{0.89} \\
    \hline
    \end{tabular}
    \caption{BMVS dataset results: comparison of RaDeGS and 2DGS methods without and with TSPE.}
    \label{table:bmvs_radegs_transposed}
  \end{minipage}\hfill
  \begin{minipage}[t]{0.48\columnwidth}
  \vspace{0pt}
    \centering
    \footnotesize
    \setlength{\tabcolsep}{2pt}
    \begin{tabular}{l|cc|cc}
    \hline
    \textbf{Scene} & \textbf{RaDe} & \textbf{\tiny{+TSPE}} & \textbf{2DGS} & \textbf{\tiny{+TSPE}} \\
    \hline
    24 & 0.40 & 0.44 & 1.58 & 0.55 \\
    37 & 0.71 & 0.74 & 1.87 & 0.82 \\
    40 & 0.33 & 0.35 & 1.55 & 0.46 \\
    55 & 0.37 & 0.41 & 1.47 & 0.51 \\
    63 & 0.87 & 0.81 & 2.43 & 0.98 \\
    65 & 0.79 & 0.74 & 1.53 & 0.86 \\
    69 & 0.77 & 0.69 & 1.34 & 0.76 \\
    83 & 1.22 & 1.22 & 1.51 & 1.19 \\
    97 & 1.26 & 1.19 & 1.57 & 1.22 \\
    105 & 0.70 & 0.63 & 1.32 & 0.87 \\
    106 & 0.65 & 0.60 & 1.26 & 0.62 \\
    110 & 0.85 & 0.88 & 2.09 & 1.26 \\
    114 & 0.33 & 0.41 & 1.42 & 0.66 \\
    118 & 0.66 & 0.66 & 1.45 & 0.66 \\
    122 & 0.44 & 0.47 & 1.17 & 0.52 \\
    \hline
    Avg$\downarrow$ & 0.68 & \textbf{0.67} & 1.57 & \textbf{0.80} \\
    \hline
    \end{tabular}
    \caption{DTU dataset results: comparison of RaDeGS and 2DGS methods without and with TSPE.}
    \label{tab:dtu_radegs_transposed}
  \end{minipage}
\end{figure}
\section{Conclusion}
We propose TSPE-GS, a probabilistic depth extraction method that reconstructs semi-transparent surfaces and occluded surfaces in Gaussian Splatting pipelines. By modeling depth distributions per pixel and progressively fusing TSDFs, our method provides a plug-and-play solution for modeling scenes with semi-transparent surfaces. Experiments show TSPE-GS outperforms previous Gaussian-based and NeRF-related methods on both semi-transparent and opaque scenes.
\textbf{Limitation.} Our method depends on the underlying 3DGS’s ability to fit scene geometry accurately; if the base 3DGS method fails to sufficiently model the geometry, the performance our approach will deteriorate.

\section*{Acknowledgements}
This work was supported by the National Science Foundation of China (62576092) and the Big Data Computing Center of Southeast University. We would like to thank anonymous reviewers for their constructive suggestions.

\nobibliography*

\bibliography{aaai2026}
\appendix
\section{Appendix A: Experimental Verification of the Unimodality of Depth Distributions}

In this appendix, we focus on verifying the unimodal nature of the depth probability density function PDF \(p(d)\) for opaque surfaces, characterized by the cumulative distribution function CDF \(P(d' < d) = 1 - T_{d'}\), where \(T_{d'}\) denotes transmittance at depth \(d'\). To this end, we employ Hartigan’s Dip Test, a non-parametric statistical test designed to assess the unimodality of empirical data.

\subsection{Experimental Procedure}

For each scene, we extract depth samples from all unmasked pixels across the BMVS dataset images, obtaining empirical cdfs of the depth. We then apply kernel density estimation (KDE) on these samples to estimate smooth approximations of the probability density functions. These estimated distributions enable modal analysis by applying Hartigan's Dip Test to statistically assess unimodality.

\subsection{Results}

Table~\ref{tab:dip_test_results} reports Hartigan's dip statistics alongside their p-values for several representative scenes. The relatively high p-values (greater than the common significance threshold of 0.05) indicate that we cannot reject the null hypothesis that the sampled depth data are unimodal. This supports our modeling assumption of unimodal depth distributions for opaque surfaces.
\begin{table}[!ht]
\centering
\footnotesize
\setlength{\tabcolsep}{8pt}
\begin{tabular}{l|c|c}
\hline
\textbf{Scene} & \textbf{Dip Statistic} & \textbf{p-value} \\
\hline
Dog       & $0.014 \pm 0.001$ & $0.27 \pm 0.12$ \\
Bear      & $0.018 \pm 0.008$ & $0.21 \pm 0.23$ \\
Durian    & $0.015 \pm 0.002$ & $0.17 \pm 0.11$ \\
Man       & $0.014 \pm 0.003$ & $0.27 \pm 0.18$ \\
Sculpture & $0.018 \pm 0.004$ & $0.08 \pm 0.07$ \\
Stone     & $0.016 \pm 0.001$ & $0.12 \pm 0.06$ \\
Clock     & $0.018 \pm 0.004$ & $0.07 \pm 0.07$ \\
\hline
Jade      & $0.025 \pm 0.003$ & $0.00 \pm 0.01$ \\
\hline
\end{tabular}
\caption{Hartigan's Dip Test Results on Depth Distributions from the BMVS Dataset}
\label{tab:dip_test_results}
\end{table}
It is worth noting that the jade scene stands out as an exception, exhibiting a significantly larger dip statistic and near-zero p-value (below 0.05), which clearly indicates a departure from unimodality. The complex geometry and porous structures in the jade scene cause its depth distribution to be multi-modal, in contrast to the relatively simple, opaque surfaces in other scenes. Therefore, our unimodal depth distribution assumption holds robustly for typical opaque objects but does not extend to such structurally intricate cases. This suggests a potential direction for future work to develop depth models capable of accommodating such complex depth distributions.

\section{Appendix B: Empirical Analysis of Multi-Peak Depth Distributions with Multiple Visible Surfaces}

This appendix provides supplementary experimental evidence supporting the observation that, in scenes containing multiple visible surfaces—including semi-transparent surfaces—the depth data can be better represented by multiple distinct depth points rather than using a single summary statistic.

\subsection{B.1 Experimental Setup}

Depth data were collected from scenes containing semi-transparent surfaces. Instead of utilizing the full depth probability distributions, we extracted depth point clouds from the peak depth values of these distributions. For comparison, point clouds generated from median depth values—which are commonly used in prior work—were also evaluated.

\subsection{B.2 Evaluation Metric}

To quantitatively compare the two point cloud sets, we use the precision metric, which measures the percentage of points in the reconstructed point cloud that lie within a small distance threshold of the ground truth surfaces. A higher precision indicates more accurate and cleaner reconstructions with fewer outliers.

\subsection{B.3 Results and Observations}

Table~\ref{tab:bottleship_precision_comparison} summarizes the precision scores of point clouds derived from peak depth versus median depth values across multiple scenes in the Bottleship dataset.

\begin{table}[!ht]
\centering
\footnotesize
\setlength{\tabcolsep}{6pt} 
\begin{tabular}{l|c|c}
\hline
\textbf{Scene} & \textbf{Prec. (Peak)} & \textbf{Prec. (Median)} \\
\hline
computer   & $0.89$ & $0.86$ \\
foodtray   & $0.91$ & $0.90$ \\
smallbuild & $0.94$ & $0.93$ \\
hotfood    & $0.85$ & $0.82$ \\
boatship   & $0.87$ & $0.81$ \\
capsule    & $0.95$ & $0.94$ \\
blood      & $0.90$ & $0.88$ \\
gumball    & $0.93$ & $0.90$ \\
\hline
\textbf{Avg.} & $\mathbf{0.90}$ & $\mathbf{0.88}$ \\
\hline
\end{tabular}
\caption{Comparison of point cloud precision (Prec.) derived from peak and median depth values on the Bottleship dataset. Precision measures the percentage of reconstructed points close to ground truth surfaces, reflecting reconstruction accuracy and quality.}
\label{tab:bottleship_precision_comparison}
\end{table}

The precision results demonstrate that point clouds extracted from peak depth values consistently achieve higher precision than those based on median depth points, indicating cleaner and more accurate reconstruction of scene geometry. This advantage holds even in the presence of noise within the underlying pixel-wise depth probability distributions. These findings confirm that leveraging multiple distinct depth peaks better captures complex scenes with multiple visible surfaces such as semi-transparent materials.

Here is a revised English version of your Appendix C content, adjusted to better match the style and technical accuracy found in your uploaded paper:

\section{Appendix C: Approximation of \(\alpha_d\) and Simplified CDF for PDF Estimation}

This appendix details the derivation of the local opacity \(\alpha_d\) approximation and introduces a simplified CDF used for efficient estimation of the depth PDF.

\subsection{C.1 Definition of Local Opacity \(\alpha_d\)}

Within a depth interval \(d_{g-1} < d < d_g\), the local opacity \(\alpha_d\) is defined via exponential interpolation as
\[
\alpha_d = 1 - \exp\left(\frac{\ln(1-\alpha_g)}{d_g - d_{g-1}} (d - d_{g-1})\right),
\]
which ensures consistency at the interval boundaries with the discrete opacity value \(\alpha_g\).

\subsection{C.2 Simplified CDF Approximation}

To facilitate efficient numerical estimation of the depth PDF \(p(d)\), the CDF \(P(d)\) that originally depends continuously on \(\alpha_d\) is instead approximated as a piecewise constant (step) function over the discrete depth partitions.

This approximation reduces computational load for sampling and PDF estimation while maintaining reasonable accuracy. In practice, instead of solving for the exact continuous depth \(d\) such that the transmittance lies between \(T_g\) and \(T_{g-1}\), the depth is approximated by the right boundary \(d_g\) of the corresponding depth interval.

\begin{figure}[ht]
    \centering
    \includegraphics[width=0.5\linewidth]{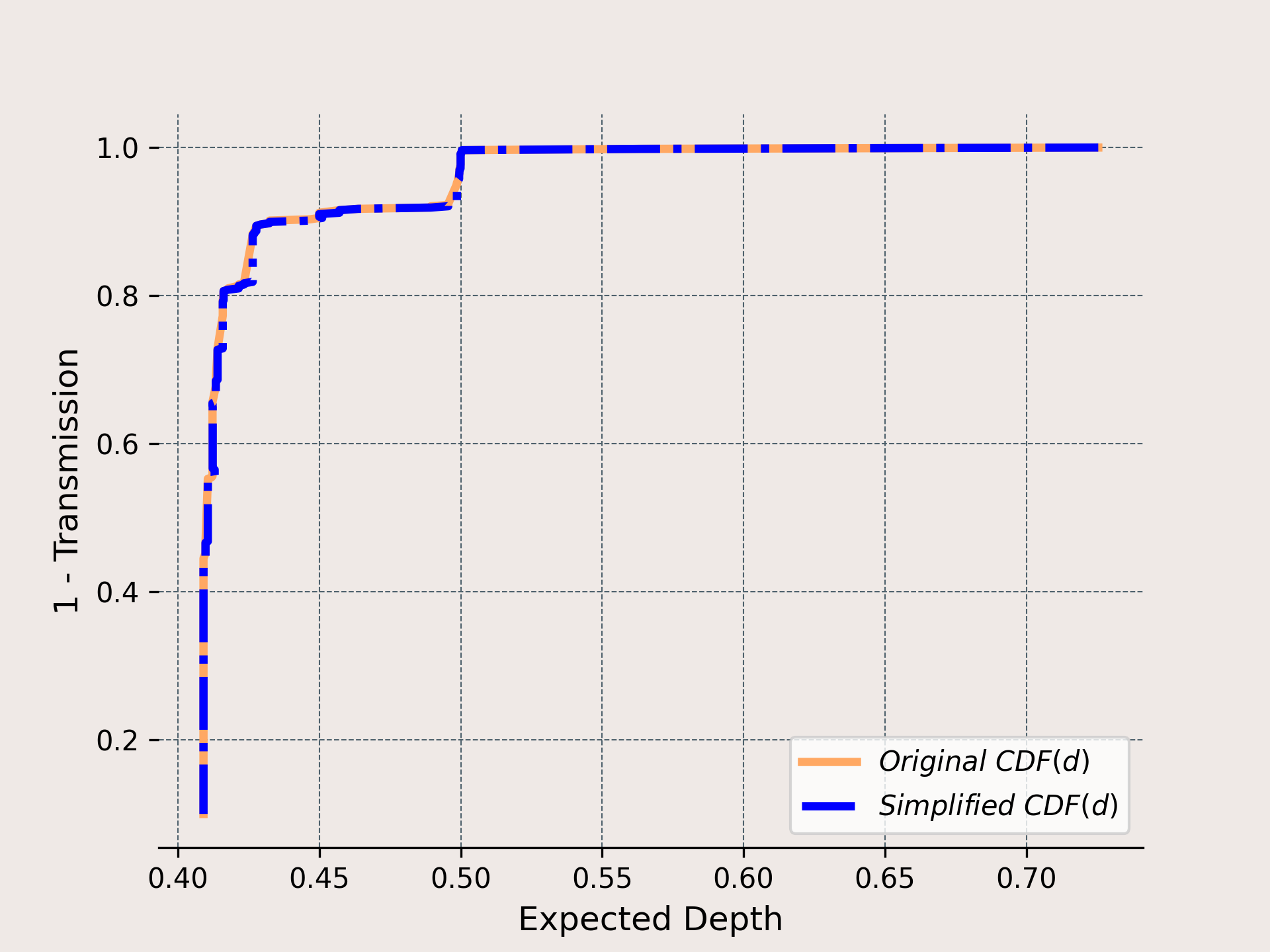}
    \caption{Comparison of CDFs derived from the continuous \(\alpha_d\) model and the simplified stepwise approximation. The curves exhibit close agreement.}
    \label{fig:cdf_comparison}
\end{figure}

\begin{figure}[ht]
    \centering
    \includegraphics[width=0.5\linewidth]{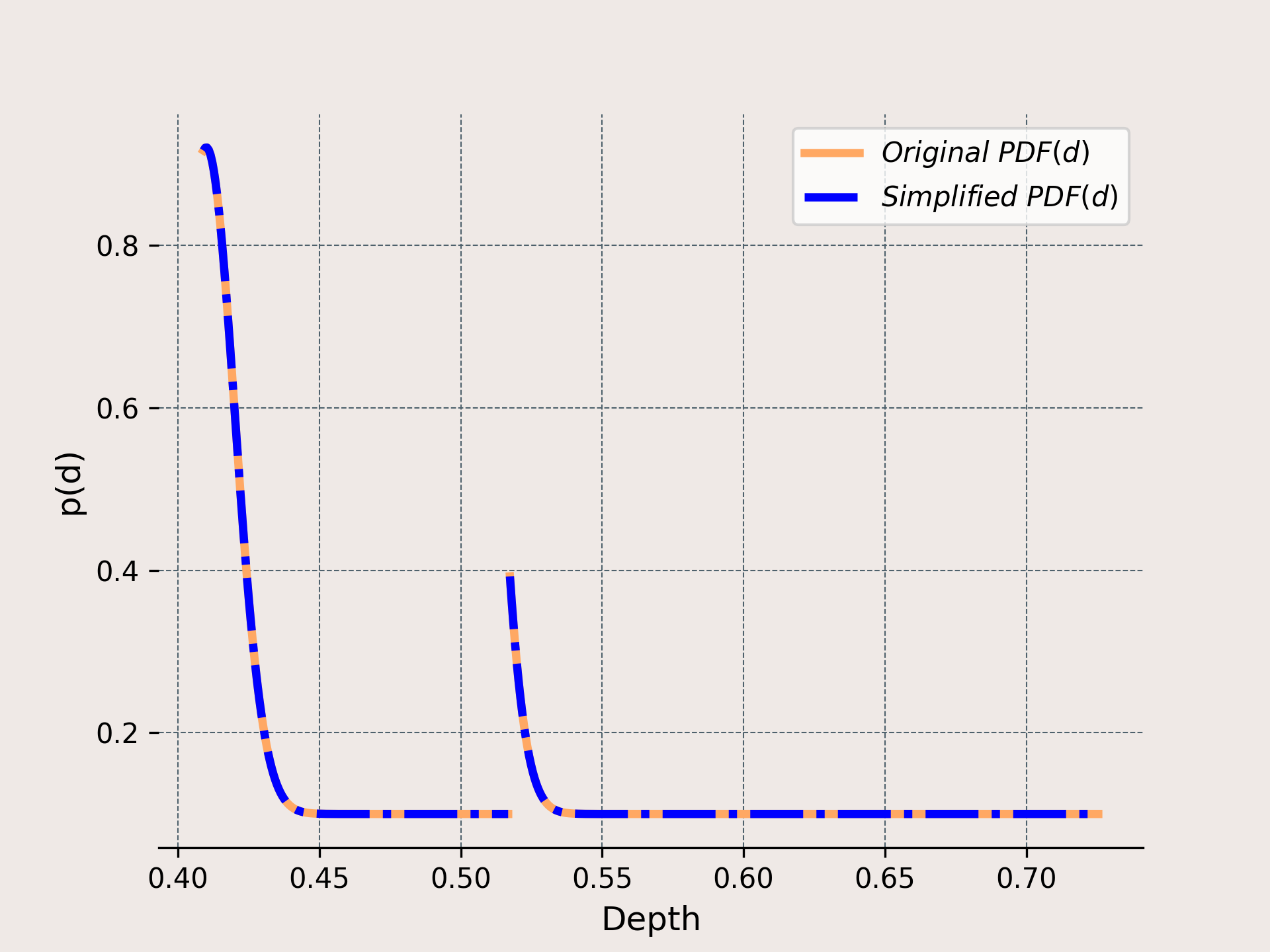}
    \caption{PDFs estimated by KDE from samples generated via the continuous \(\alpha_d\)-based model and the simplified CDF approximation. The curves nearly coincide, indicating the approximation is a valid substitute.}
    \label{fig:pdf_comparison}
\end{figure}

To validate the proposed simplified CDF approximation, Figure~\ref{fig:cdf_comparison} compares the continuous CDF derived from \(\alpha_d\) with the stepwise approximation, showing a close match. Complementarily, Figure~\ref{fig:pdf_comparison} plots KDE estimates of PDFs generated from samples using both methods, exhibiting nearly identical curves. These figures are generated based on data from the Bottleship dataset, demonstrating that the approximation maintains accuracy without sacrificing computational efficiency in practical scenarios.

\subsection{C.3 Experimental Validation and Summary}

Experimental results confirm that the simplified CDF approximation produces PDF estimates closely matching those from the original continuous model. The small discrepancies have negligible effects on downstream applications such as expected depth calculation and probabilistic modeling.

This approximation thus provides a computationally efficient yet accurate approach for multi-layer depth PDF estimation, suitable for practical implementations.
\section{Appendix D: Analysis of Multi-Layer Depth Fusion Artifacts in TSDF Integration}
\subsection{D.1 Overview of Multi-Layer TSDF Fusion}

When dealing with scenes that include transparent or semi-transparent surfaces, a single camera ray can penetrate multiple surfaces, generating multi-layer depth information. When these multi-layer depth maps, acquired from different viewpoints (e.g., Camera A and Camera B), are integrated into the same Truncated Signed Distance Function (TSDF) volume, inconsistencies can arise between the surfaces reconstructed from each view, leading to fusion artifacts.

\subsection{D.2 Example: Transparent Object with an Internal Object from a Dual-Camera View}

\begin{figure}[h!]
    \centering
    \includegraphics[width=0.3\textwidth]{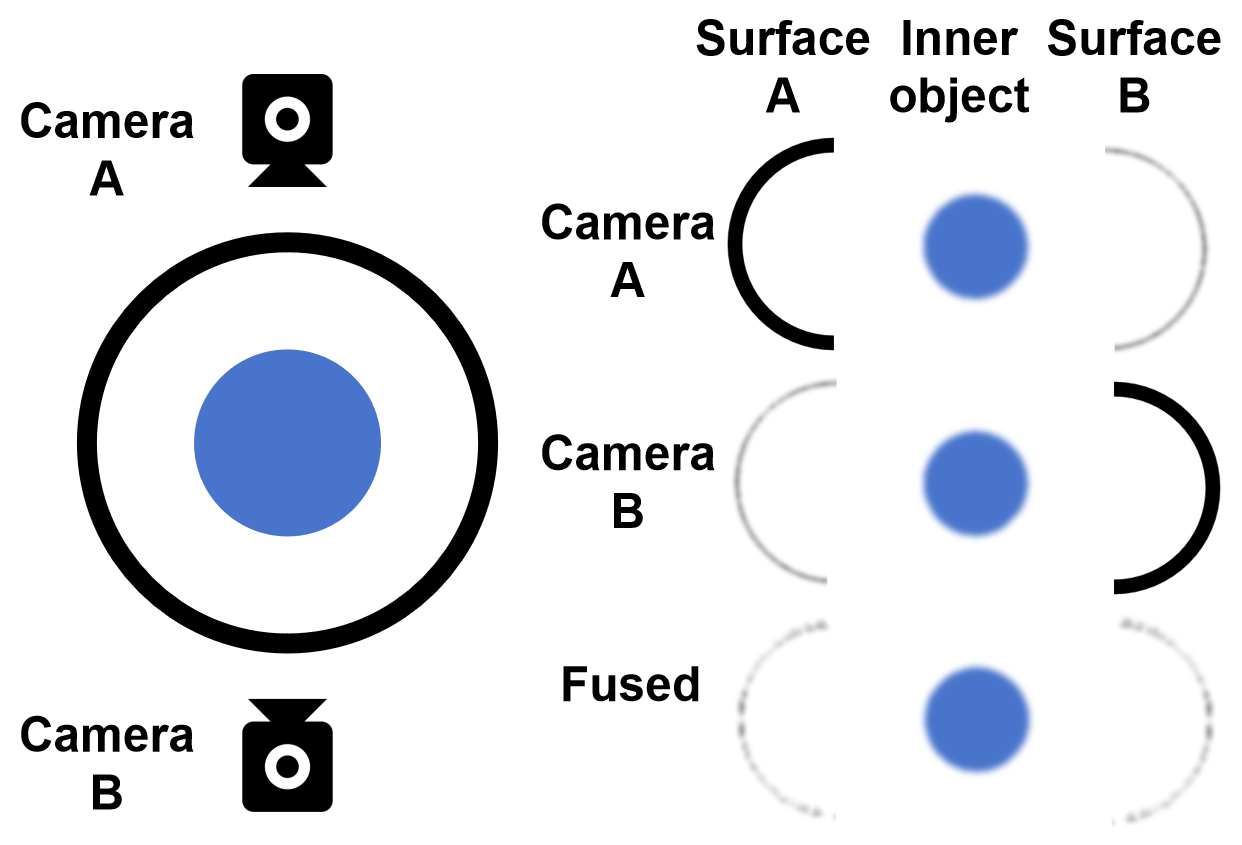}
    \caption{Schematic diagram illustrating the cause of fusion artifacts. The left panel shows the physical setup with two cameras viewing a transparent object with an internal component. The right panel shows the individual surfaces reconstructed by each camera and the resulting ``ghost'' artifact when these conflicting views are fused.}
    \label{fig:fusion_artifact_diagram}
\end{figure}

To illustrate this problem, let's consider the scenario depicted in Figure~\ref{fig:fusion_artifact_diagram}. The scene consists of a hollow, circular transparent object (like a cup) with an opaque object inside. Two cameras, Camera A and Camera B, observe this scene from opposite directions.

As shown on the left side of Figure~\ref{fig:fusion_artifact_diagram}, Cameras A and B are positioned at the top and bottom, respectively, both aimed at the central transparent object and the blue object within it. The right side of the figure details how each camera perceives the scene and the result of fusing their views.

\begin{itemize}
    \item \textbf{Camera A's View:} Observing from the top, Camera A clearly captures the outer surface of the transparent object closer to it (Surface A) and the blue object inside. However, its perception of the other side of the transparent object, Surface B, is less distinct and blurry.
    \item \textbf{Camera B's View:} Similarly, observing from the bottom, Camera B clearly reconstructs the outer surface it faces (Surface B) and the internal object. However, its reconstruction of the opposite surface, Surface A, is poor.
    \item \textbf{Fused Result:} When these two views are fused into a single TSDF model, the internal blue object is reconstructed accurately because it was clearly captured by both cameras. However, the situation becomes more complex for the outer transparent circular object. Because Camera A has a blurry view of Surface B and Camera B has a blurry view of Surface A, forcing the fusion of this uncertain and conflicting information prevents the formation of a clear, complete surface. Ultimately, the outer circular object is reconstructed as a faint, discontinuous ``ghost'' shape---a classic example of a fusion artifact.
\end{itemize}

\begin{figure}[h!]
    \centering
    \includegraphics[width=0.9\linewidth]{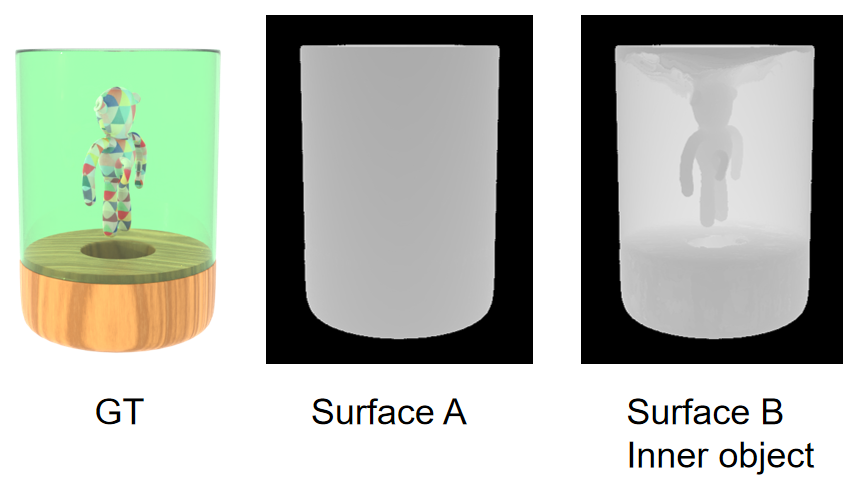}
    \caption{Visualization of multi-layer surface reconstruction by Camera A in the real scene.}
    \label{fig:multi_layer_surface}
\end{figure}

To further illustrate the reconstruction challenges from a single viewpoint and highlight the multi-layer depth complexity, Figure~\ref{fig:multi_layer_surface} shows the actual surfaces reconstructed by Camera A in a real scene. The left panel depicts the GT model of the transparent container with the internal object, serving as the reference. The middle panel shows the outer surface (Surface A) which Camera A observes directly and reconstructs with reasonable clarity. The right panel displays the inner surface (Surface B) along with the internal object, as perceived by Camera A through the transparent material but with less definition and accuracy due to occlusions and optical distortions. This data corresponds to the ``monkey'' scene from the $\alpha$Surf dataset.

\subsection{D.3 Cause of Fusion Artifacts and Conclusion}

This case clearly demonstrates the inherent inconsistencies in the reconstruction of ``inner'' or ``occlude'' layers from different viewpoints during multi-layer depth fusion. As illustrated in Figure~\ref{fig:fusion_artifact_diagram}, when this conflicting depth information is directly integrated through weighted averaging, the TSDF voxels receive contradictory update commands. This leads to the final reconstructed surface suffering from defects such as holes, depressions, roughness, and even ``ghost'' artifacts.

This example highlights the limitations of directly applying traditional TSDF fusion methods to multi-layer depth maps, especially when handling transparent or complex layered scenes. Therefore, developing advanced fusion algorithms that can better handle probabilistic multi-layer depth data and effectively identify and resolve inter-view conflicts is crucial for high-fidelity 3D reconstruction.

\section{ Appendix E: Progressive Reconstruction Strategy and Algorithm}

This appendix details the proposed progressive TSDF fusion strategy designed to mitigate interference issues when fusing multi-layer depth maps extracted from probabilistic distributions.

\subsection{Motivation}

Directly fusing all peak depth maps into a TSDF volume often causes surface artifacts such as roughness and holes due to interference between overlapping inner and outer layers. These problems mainly stem from less accurate inner-layer depths corrupting the outer surface reconstruction, as mentioned in Appendix D.

\subsection{Proposed Progressive TSDF Fusion}

To address this, we introduce a two-stage fusion algorithm:

1. Outer Layer Fusion: Integrate outermost depth maps first and freeze the voxels with significant updates to protect the outer surface integrity.

2. Inner Layer Fusion: Subsequently integrate inner layer depth maps while skipping voxels already frozen, preventing interference with outer surfaces.

\subsection{Algorithm}
The following pseudocode presents our progressive TSDF fusion algorithm for multi-layer depth map integration:
\begin{algorithm}[!ht]
\caption{Progressive TSDF Fusion}
\label{alg:progressive_tsdf_fusion}
\textbf{Input}:  
Outer layer depth maps \(\{D^{outer}_i\}\), inner layer depth maps \(\{D^{inner}_j\}\), camera poses \(\{P_k\}\), initial TSDF \(V\) \\
\textbf{Output}:  
Fused TSDF volume \(V\)

\begin{algorithmic}[1]
\STATE Initialize the TSDF volume \(V\) and weight volume \(W\)
\STATE Initialize frozen voxel set \(F \leftarrow \emptyset\)

\FOR{each outer depth map \(D^{outer}_i\) with camera pose \(P_i\)}
  \FOR{each voxel \(v \in V\) intersected by camera ray from \(P_i\)}
    \STATE Compute TSDF update based on \(D^{outer}_i\) and update \(V[v]\), \(W[v]\)
    \IF{weight update for voxel \(v\) exceeds threshold}
      \STATE Add voxel \(v\) to frozen set \(F\)
    \ENDIF
  \ENDFOR
\ENDFOR

\FOR{each inner depth map \(D^{inner}_j\) with camera pose \(P_j\)}
  \FOR{each voxel \(v \in V\) intersected by camera ray from \(P_j\)}
    \IF{voxel \(v \notin F\)}
      \STATE Compute TSDF update based on \(D^{inner}_j\) and update \(V[v]\), \(W[v]\)
    \ENDIF
  \ENDFOR
\ENDFOR

\STATE \textbf{return} the fused TSDF volume \(V\)
\end{algorithmic}
\end{algorithm}

\section{Appendix F: Analysis of Reconstruction Results and Method Compatibility}

In this appendix, we present a qualitative analysis by visualizing the distributions of Gaussian primitives, which effectively represent the expressive capacity of 3DGS methods in modeling scene geometry. The data for semi-supervised scenes used in these analyses are sourced from the Bottleship dataset.

\begin{figure}[ht]
    \centering
    \includegraphics[width=0.9\linewidth]{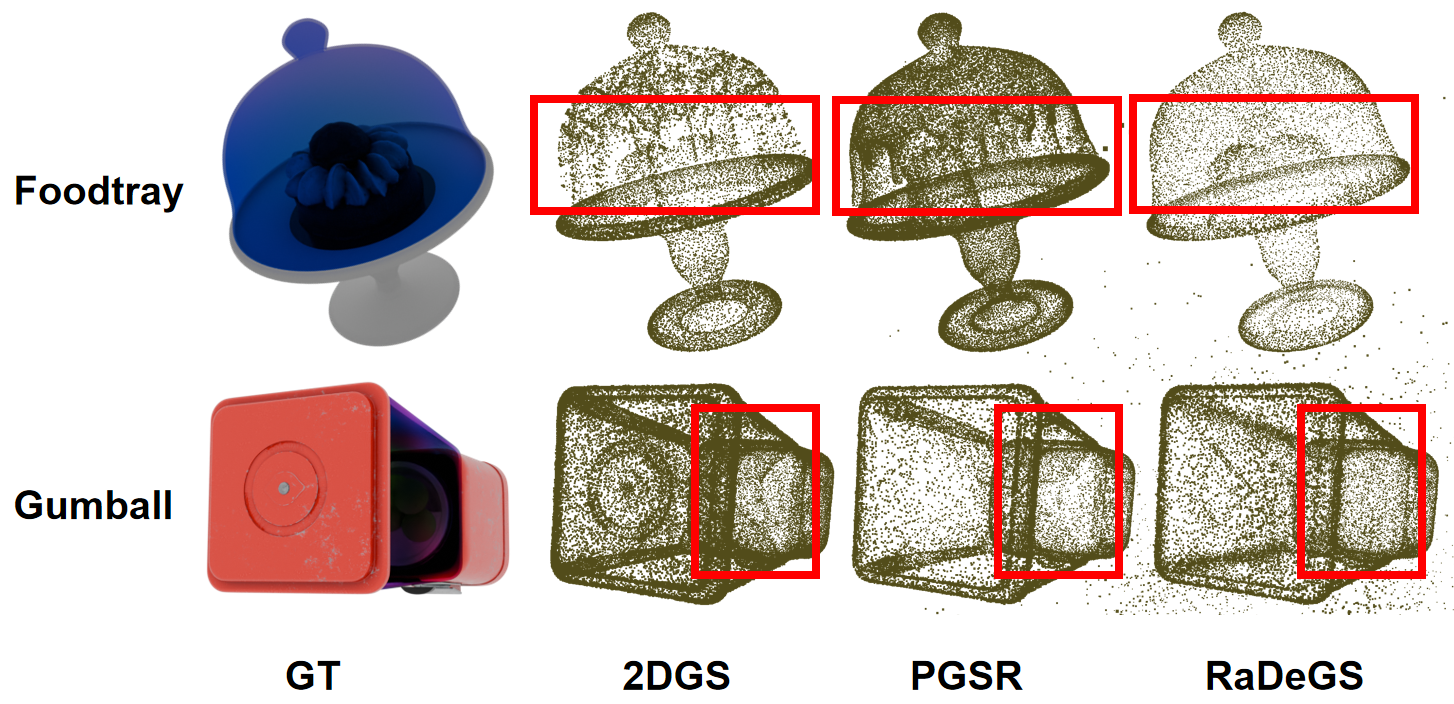}
    \caption{Visualization of Gaussian primitive distributions from 2DGS, PGSR, and RaDeGS on semi-transparent objects (``Foodtray'' and ``Gumball''), illustrating differences in spatial coverage and artifact presence on transparent surfaces.}
    \label{fig:gaussian_distributions}
\end{figure}

Figure~\ref{fig:gaussian_distributions} shows the spatial distributions of Gaussian primitives generated by different methods on semi-transparent objects. We observe that PGSR often produces noticeable concave artifacts on transparent regions, indicating less accurate surface coverage. In contrast, 2DGS tends to exhibit slab-like ghosting artifacts, appearing as planar discontinuities that degrade reconstruction quality. Compared to these methods, RaDeGS demonstrates a more stable and consistent distribution of Gaussian primitives across transparent surfaces, which enables more reliable and accurate reconstructions.

Based on these observations, although our analysis remains qualitative, we find that RaDeGS provides better stability and coverage in representing complex, semi-transparent geometries. Accordingly, we adopt RaDeGS as our foundational method in this work.

\end{document}